\documentclass[10pt,twocolumn,letterpaper]{article}

\usepackage[usenames,dvipsnames]{xcolor}

\usepackage{misc/iccv}
\usepackage{times}
\usepackage{epsfig}
\usepackage{graphicx}
\usepackage{amsmath}
\usepackage{amssymb}

\usepackage{algorithmic}
\usepackage{algorithm}
\usepackage{rotating}
\usepackage{diagbox}
\usepackage{balance}

\usepackage[utf8]{inputenc} 
\usepackage[T1]{fontenc}    
\usepackage{url}            
\usepackage{booktabs}       
\usepackage{amsfonts}       
\usepackage{amsmath}
\usepackage{nicefrac}       
\usepackage{microtype}      
\usepackage{graphicx}       

\usepackage{paralist}
\usepackage{xspace}
\usepackage{comment}
\usepackage{subfig}

\newcommand{\myparagraph}[1]{\vspace{0.1em}\noindent\textbf{#1}}

\newcommand{\figref}[1]{Fig.~\ref{#1}}
\newcommand{\norm}[1]{\left\lVert#1\right\rVert}

\usepackage[pagebackref=true,breaklinks=true,letterpaper=true,colorlinks,bookmarks=false]{hyperref}

\iccvfinalcopy

\ificcvfinal\fi
\begin{document}

\title{Monocular Neural Image Based Rendering with Continuous View Control}

\author{Xu Chen\footnotemark[1] \quad \quad Jie Song\footnotemark[1] \quad \quad Otmar Hilliges \vspace{0.1cm} \\
AIT Lab, ETH Zurich\\
{\tt\small \{xuchen, jsong, otmar.hilliges\}@inf.ethz.ch}
}

\newcommand{\figureNetwork}
{
\begin{figure*}
\centering
    \includegraphics[trim={0 220 0 0},clip,width=1\linewidth]{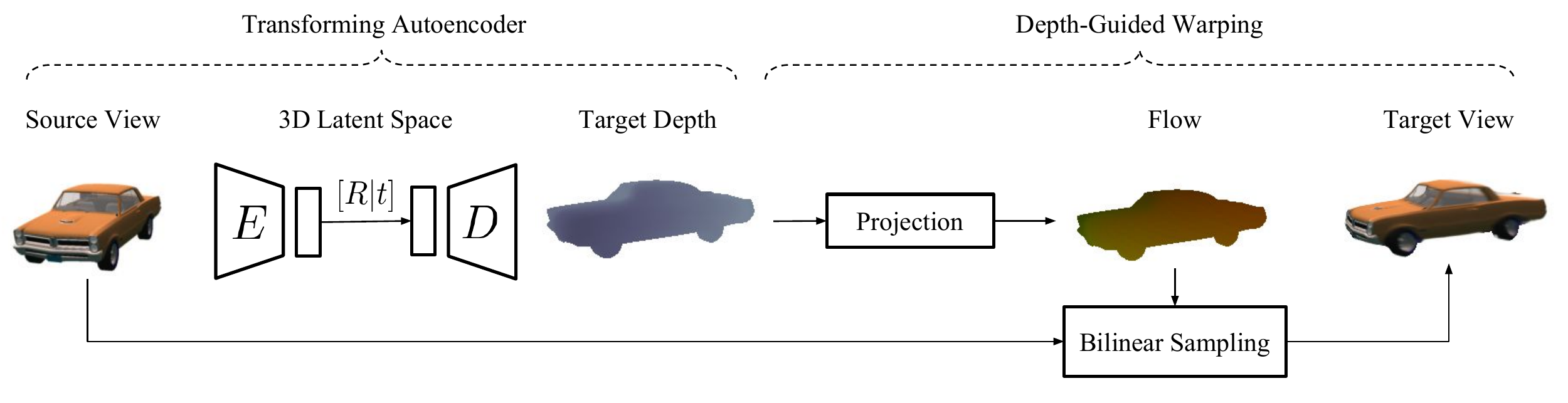}
    \vspace{-0.7cm}
    \caption{
    \textbf{Pipeline overview.}
    2D source views are encoded and the latent code is explicitly rotated before a decoder network predicts the depth map in the target view. 
    Dense correspondences are attained via perspective projection and used to warp pixels from source view to the target with bilinear sampling.
    All operations are differentiable and trained end-to-end without ground-truth depth or flow maps. 
    The only supervision is a $L_1$ reconstruction loss between target view and ground truth image.
    }
    \label{fig:pipeline}
\end{figure*}
}

\newcommand{\figureShapeNet}
{
\begin{figure*}[!t]
    \includegraphics[width=1\textwidth]{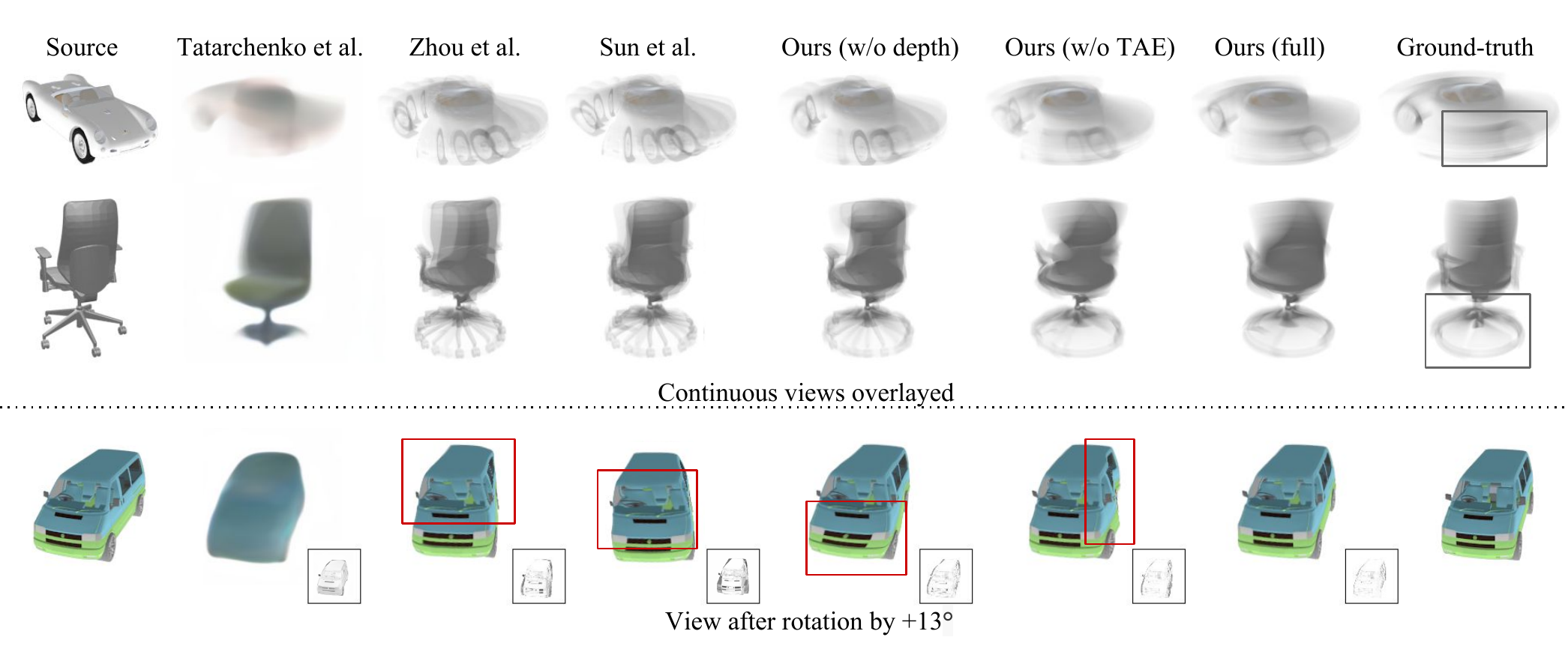}
    \caption{
    \textbf{Qualitative results for granularity and precision of viewpoint control on ShapeNet.}
    In the top two rows, we generate and overlay 80 continuous views with step size of 1$^\circ$ from a single input.
    Our method exhibits similar spin pattern as the ground truth, whereas other methods mostly converge to the fixed training views (see wheels of the car and chair indicated by the box). 
    In the bottom row, a close look at specific views is given, which reveals that previous methods display distortions or converge to neighboring training views (Zhou et al.\cite{zhou2016view}, Sun et al.\cite{sun2018multiview}). The image generated by Tatarchenko et al.\cite{TDB16a} is blurry. Corresponding error maps are also depicted. 
    \emph{Best viewed in color.}
    }
    \label{fig:large_shapenet}
\end{figure*}
}

\newcommand{\figureControl}
{
\begin{figure}[h]
    \includegraphics[trim={25 30 40 65},clip,width=.95\linewidth]{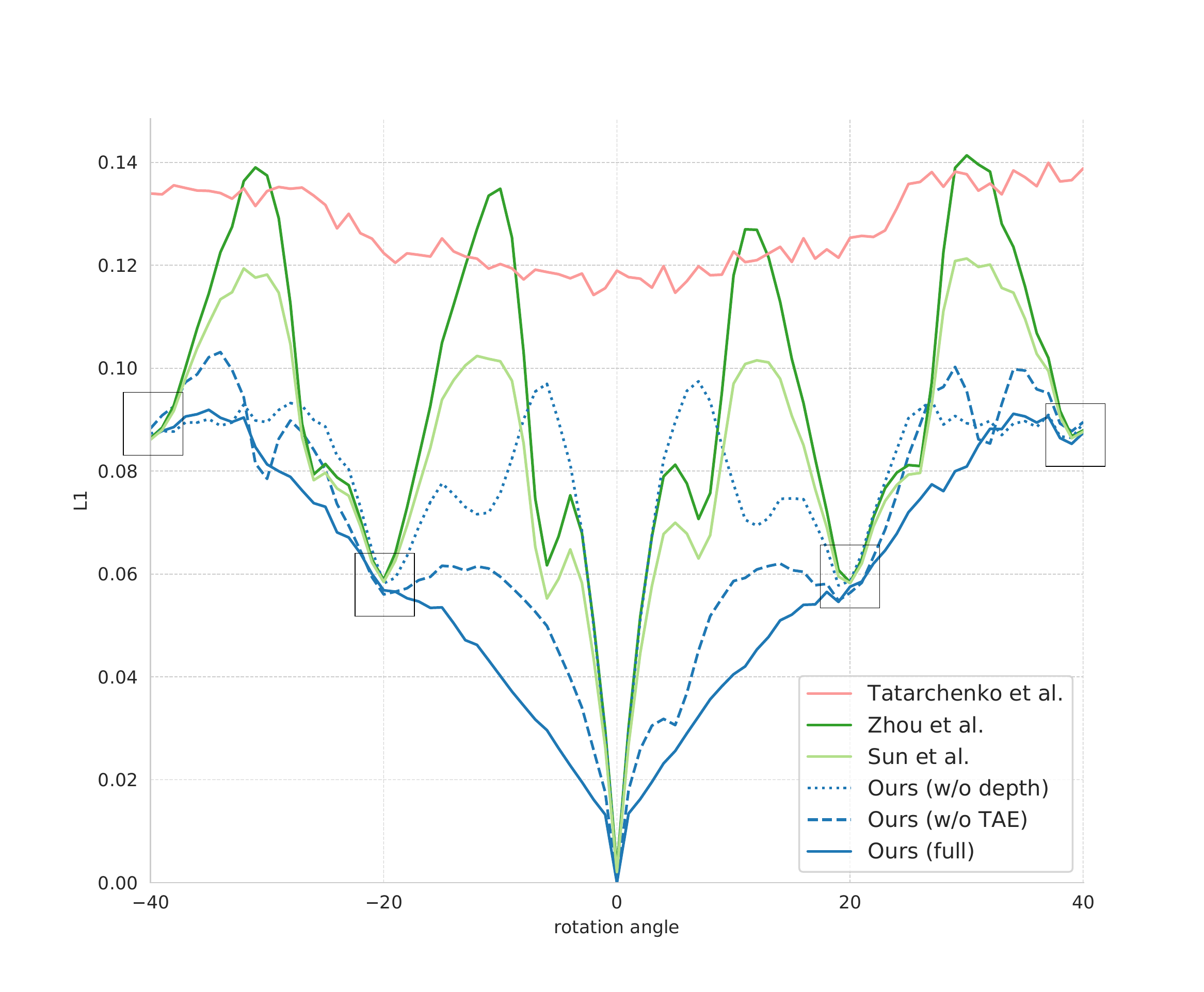}
    \caption{
    Comparison of $L_1$ \textbf{reconstruction error} as a function of view rotation on \textbf{car}.
    Ours outperforms other state-of-the-art baselines over the entire range and yields a smoother loss progression.  Note that $0^\circ$ here means no transformation applied to the source view. ($\pm40^\circ,\pm20^\circ$ are training views indicated by black boxes).
    }
    \label{fig:control}
\end{figure}
}

\newcommand{\figureResultKITTISLAM}
{
\begin{figure*}[!t]
    \includegraphics[width=1\textwidth]{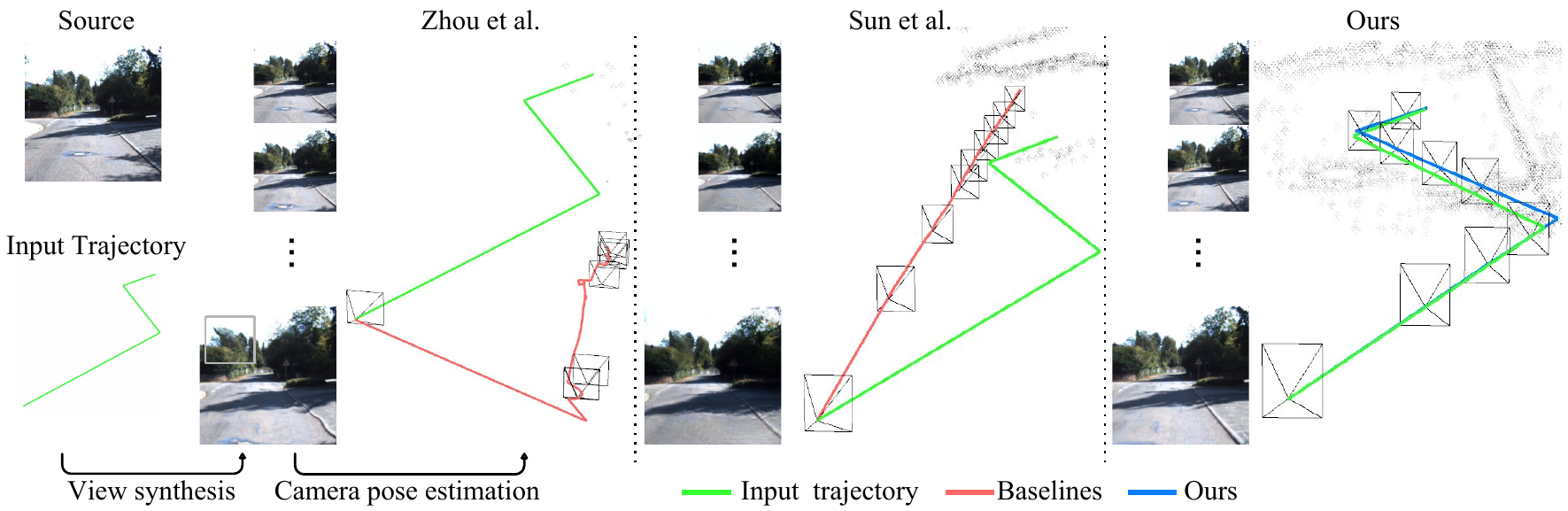}
    \caption{
    \textbf{Complex camera motion.} 
    Setting: given a source view and an input trajectory, a continuous sequence of views is synthesized along the user defined trajectory (green). 
    Trajectories are estimated via a state-of-the-art visual odometry system \cite{engel2017direct} and compared to the desired trajectory. 
    The trajectory estimated from \textit{Ours} align well with the ground-truth, while \cite{zhou2016view,sun2018multiview} mostly produce straight forward or wrong motion regardless of the input.
    }
    \label{fig:large_kitti}
\end{figure*}
}

\newcommand{\figureResultKITTIImage}
{
\begin{figure*}[!t]
    \includegraphics[width=1\textwidth]{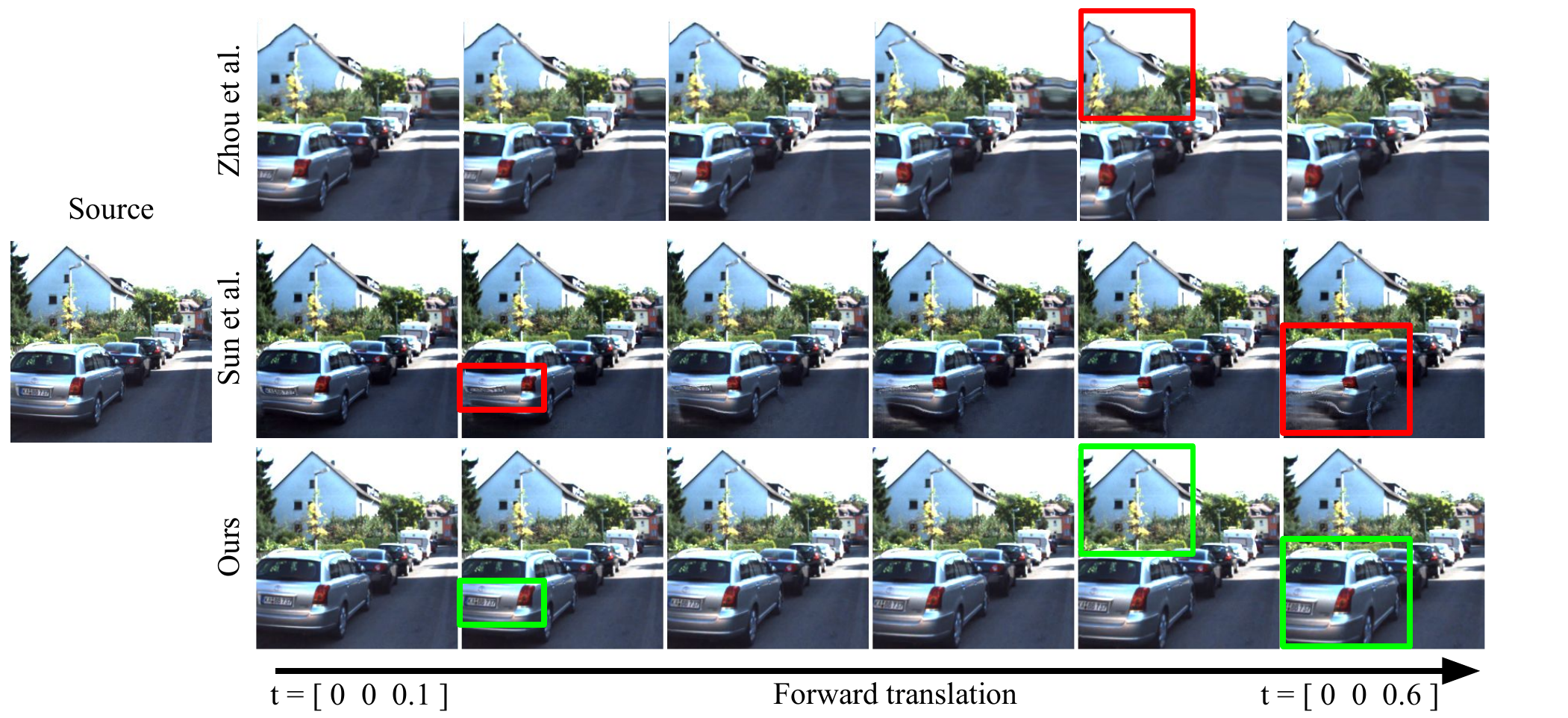}
    \caption{
    \textbf{Simple camera motion.} 
    Setting: Given a source view we synthesize linear forward motion over 0.6m.
    Our method produce sharp and correct images while \cite{zhou2016view,sun2018multiview} produces distorted images. Zhou et al.~\cite{zhou2016view}'s motion is incorrect, while Sun et al.~\cite{sun2018multiview} stays stationary. Ours reflects a reasonable straight forward transition.
    }
    \label{fig:large_kitti2}
\end{figure*}
}

\newcommand{\tableShapeNetView}
{
\begin{table}[h]
\centering
\begin{tabular}{c|cc|cc}
                                           & \multicolumn{2}{c|}{Car}                     & \multicolumn{2}{c}{Chair}                  \\
\multicolumn{1}{l|}{}                      & L1                   & SSIM                  & L1                   & SSIM                 \\ \hline
Tatarchenko et al. \cite{TDB16a}                             & 0.084                & 0.919                 & 0.110                & 0.917                \\
Zhou et al. \cite{zhou2016view}                             & 0.062                & 0.924                 & 0.074                & 0.920                \\
Sun et al. \cite{sun2018multiview}                             & 0.056                & 0.926                 & 0.070                & 0.921                \\
Ours (w/o depth)                           & 0.052                & 0.932                 & 0.066                & 0.926                \\
Ours (w/o TAE)                              & 0.045                & 0.943                 & 0.065                & 0.930                \\
Ours (full)                                & \textbf{0.039}                & \textbf{0.949}                 & \textbf{0.056}                & \textbf{0.938}               
\end{tabular}
\caption{\textbf{Quantitative analysis of fine-grained view control on ShapeNet.} Average L1 error and SSIM for all generated views between $[-40^\circ,40^\circ]$ from the source view.}
\label{tab:shapenet_view}
\end{table}
}

\newcommand{\tableDepthFlow}
{
\begin{table}[h]
\centering
\begin{tabular}{c|cc|cc}
& \multicolumn{2}{c|}{Flow} & \multicolumn{2}{c}{Depth} \\
\multicolumn{1}{l|}{} & L1          & Acc         & L1          & Acc          \\ \hline
Zhou et al. \cite{zhou2016view}   & 0.035       & 69.1\%      & -           & -            \\
Ours (w/o depth)      & 0.029       & 76.3\%      & -           & -            \\
Ours (w/o TAE)         & 0.022       & 84.6\%      & 0.134       & 89.0\%       \\
    Ours (full)           & \textbf{0.021}       & \textbf{85.7\%}      & \textbf{0.132}       & \textbf{91.1\%}      
\end{tabular}
\caption{\textbf{Quantitative analysis of flow and depth prediction on car.} Average $L_1$ error and accuracy for all predicted flow and depth in target views between $[-40^\circ,40^\circ]$ from the source view. Ours significantly outperforms the baselines.}
\label{tab:depth_flow}
\end{table}
}

\newcommand{\tableKITTIView}{
\begin{table}[h]
\centering

\begin{tabular}{c|cc}
\multicolumn{1}{l|}{} & TE                   & RE                   \\ \hline
Zhou et al. \cite{zhou2016view}        & 0.557                & 0.086                \\
Sun et al. \cite{sun2018multiview}        & 0.435  & 0.080  \\ 
Ours           & \textbf{0.108}                & \textbf{0.019}               
\end{tabular}
\caption{\textbf{Precision evaluation} of viewpoint control by camera pose estimation on KIITI.}
\label{tab:kitti_view}
\end{table}}

\newcommand{\figurePCL}
{
\begin{figure}[h]
    \includegraphics[width=\linewidth]{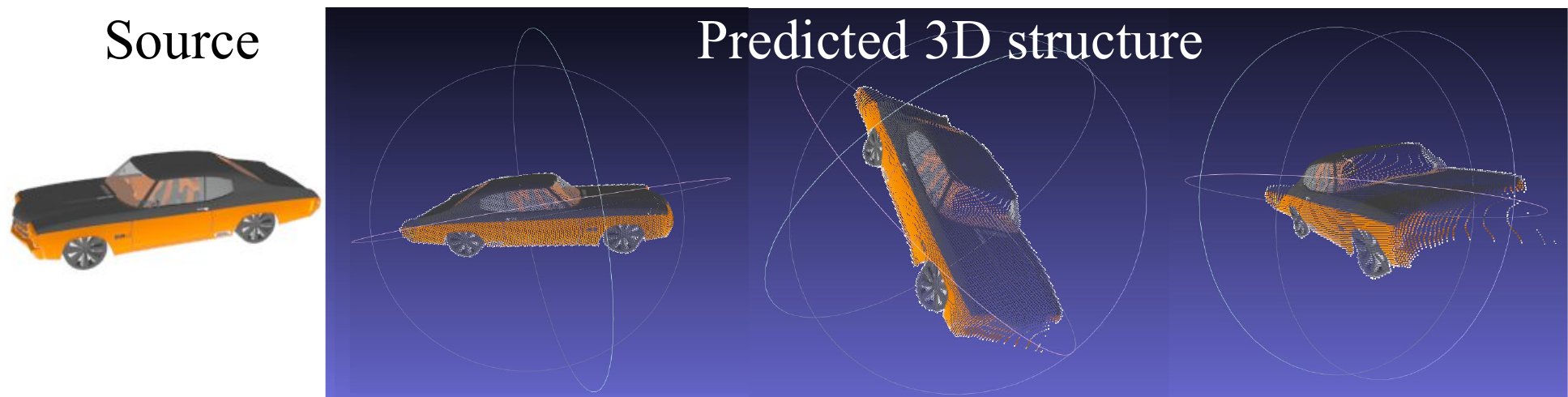}
    \caption{
    \textbf{Unsupervised depth prediction.}
     Depth map is predicted from the source view and visualized as point clouds depicted from different viewing angles. 
    } 
    \label{fig:pcl}

\end{figure}
}

\newcommand{\figureInterpolation}{
\begin{figure}[!h]
\centering
    \includegraphics[width=\linewidth] {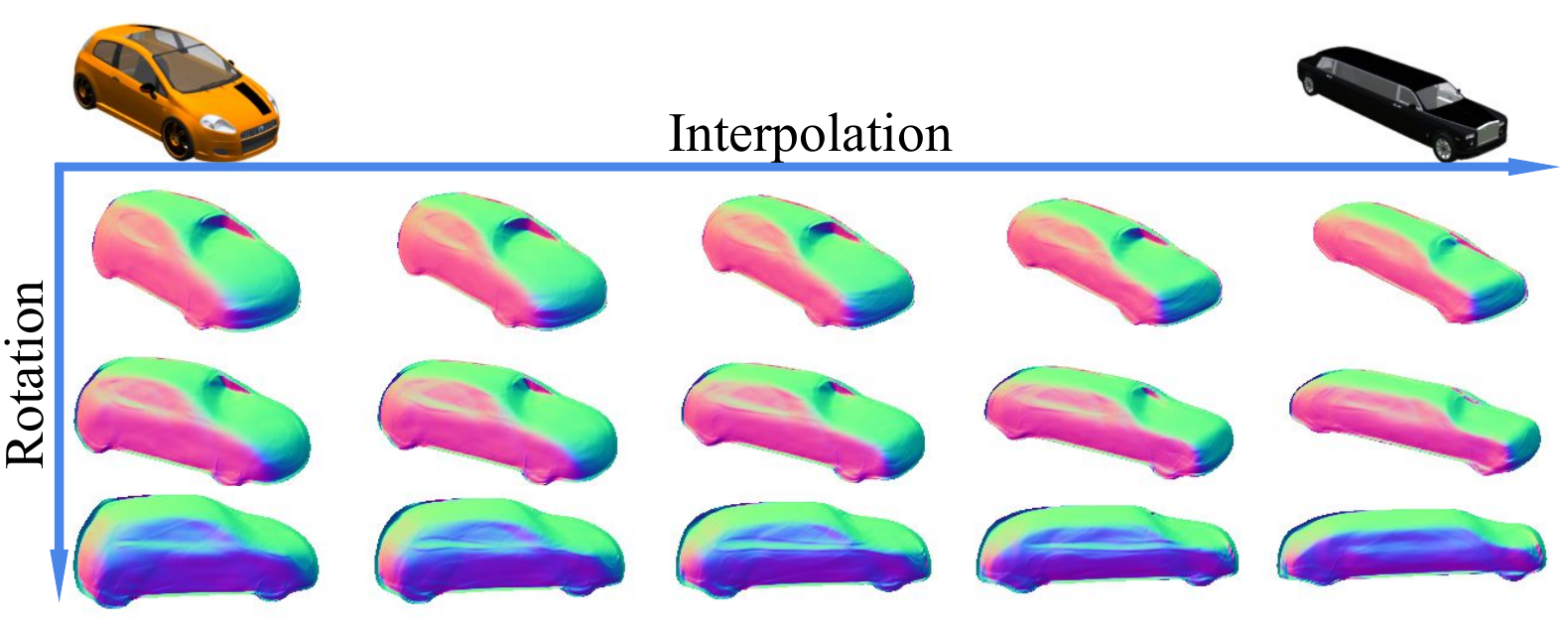}
    \caption{
    \textbf{Latent space analysis showing consistency of embeddings}. Left-to-right: latent space interpolation between \emph{different} objects. Top-to-bottom: Rotation of \emph{same} latent code. (Normals in global frame, extracted from depth). }
    \label{fig:inter}
\end{figure}
}

\twocolumn[{%
\renewcommand\twocolumn[1][]{#1}%
\vspace{-0.6cm}
\maketitle
\begin{center}
    \vspace{-1cm}
    \includegraphics[trim={0 0 0 0},clip,width=1\linewidth]{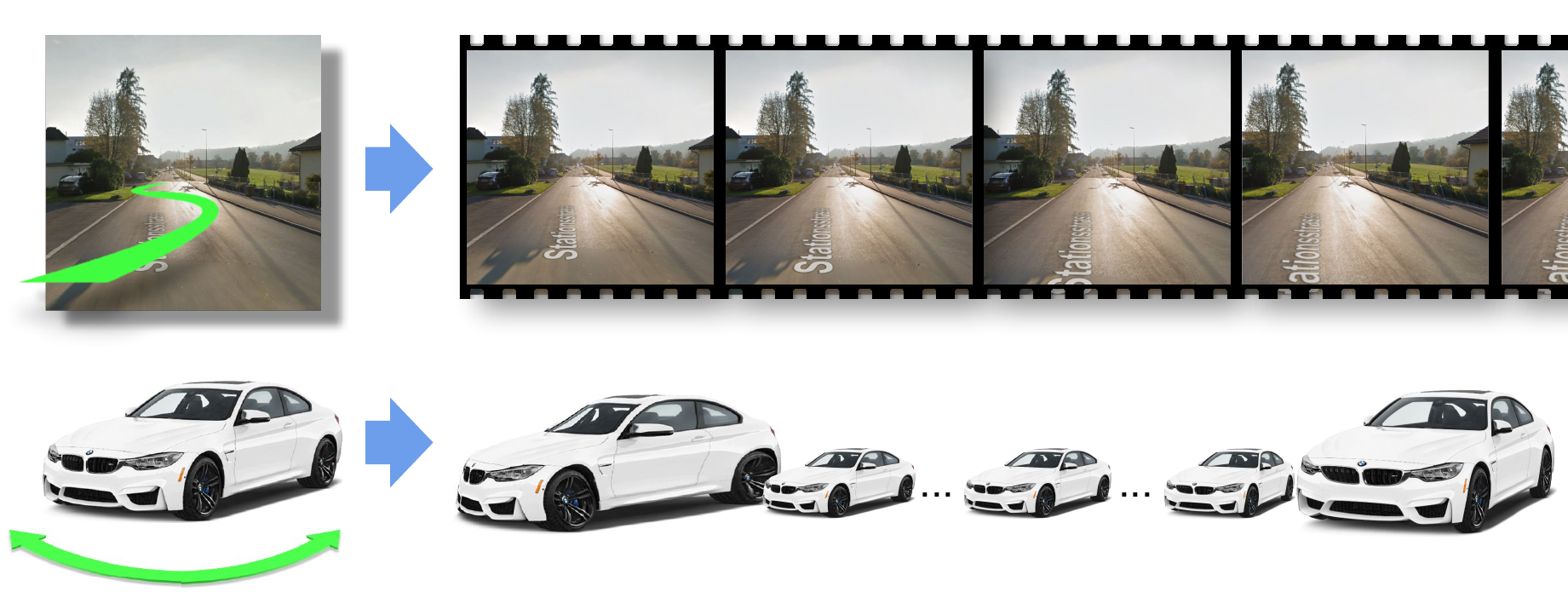}
    \vspace{-0.6cm}
    \captionof{figure}{
    \textbf{Interactive novel view synthesis:} 
    Given a single source view our approach can generate a continuous sequence of geometrically accurate novel views under fine-grained control. 
    \emph{Top}: Given a single street-view like input, a user may specify a continuous camera trajectory and our system generates the corresponding views in real-time. 
    \emph{Bottom}: An unseen hi-res internet image is used to synthesize novel views, while the camera is controlled interactively. 
    \emph{Please refer to our project homepage{\textsuperscript{$\dagger$}}.}
    }
    \label{fig:teaser}
\end{center}
}
]

\footnotetext[1]{Equal contribution.}

\begin{abstract}
\vspace{-0.3cm}
We propose a method to produce a continuous stream of novel views under fine-grained (e.g., 1$^\circ$ step-size) camera control at interactive rates. 
A novel learning pipeline determines the output pixels directly from the source color. Injecting geometric transformations, including perspective projection, 3D rotation and translation into the network forces implicit reasoning about the underlying geometry. 
The latent 3D geometry representation is compact and meaningful under 3D transformation, being able to produce geometrically accurate views for both single objects and natural scenes.
Our experiments show that both proposed components, the transforming encoder-decoder and depth-guided appearance mapping, lead to significantly improved generalization beyond the training views and in consequence to more accurate view synthesis under continuous 6-DoF camera control. Finally, we show that our method outperforms state-of-the-art baseline methods on public datasets.
\end{abstract}


\vspace{-0.5cm}
\section{Introduction}

3D immersive experiences can benefit many application scenarios. For example, in an online store one would often like to view products interactively in 3D rather than from discrete view angles. Likewise in map applications it is desirable to explore the vicinity of street-view like images beyond the position at which the photograph was taken.
This is often not possible because either only 2D imagery exists, or because storing and rendering of full 3D information does not scale. 
To overcome this limitation we study the problem of interactive view synthesis with 6-DoF view control, taking only a single image as input. 
We propose a method that can produce a continuous stream of novel views under fine-grained (e.g., 1$^\circ$ step-size) camera control (see \figref{fig:teaser}).

\footnotetext[2]{https://ait.ethz.ch/projects/2019/cont-view-synth/}

Producing a continuous stream of novel views in \emph{real-time} is a challenging task. 
To be able to synthesize high-quality images one needs to reason about the underlying geometry. However, with only a monocular image as input the task of 3D reconstruction is severely ill-posed. 
Traditional image-based rendering techniques do not apply to the real-time monocular setting since they rely on multiple input views and also can be computationally expensive. 

Recent work has demonstrated the potential of learning to predict novel views from monocular inputs by leveraging a training set of viewpoint pairs \cite{TDB16a,zhou2016view,tvsn_cvpr2017,dosovitskiy2017learning}.
This is achieved either by directly synthesizing the pixels in the target view~\cite{TDB16a,dosovitskiy2017learning} or predicting flow maps to warp the input pixels to the output~\cite{zhou2016view,sun2018multiview}.
However, we experimentally show that such approaches are prone to over-fitting to the training views and do not generalize well to free-from non-training viewpoints. 
If the camera is moved continuously in small increments, with such methods, the image quality quickly degrades. One possible solution is to incorporate much denser training pairs but this is not practical for many real applications.
Explicit integration of geometry representations such as meshes~\cite{kato2018renderer,DBLP:journals/corr/abs-1901-05567} or voxel grids~\cite{wu20153d,choy20163d,girdhar2016learning,tulsiani2017multi} could be leveraged for view synthesis. However, such representations would limit applicability to settings where the camera orbits a single object. 

In this paper, we propose a novel learning pipeline that determines the output pixels directly from the source color but forces the network to implicitly reason about the underlying geometry. 
This is achieved by injecting geometric transformations, including perspective projection, 3D rotations and translations into an end-to-end trainable network. The latent 3D geometry representation is compact and memory efficient, is meaningful under explicit 3D transformation and can be used to produce geometrically accurate views for both single objects and natural scenes.
More specifically, we propose a geometry aware neural architecture consisting of a 3D transforming autoencoder (TAE) network~\cite{Hinton:2011:TA:2029556.2029562} and subsequent depth-guided appearance warping. 
In contrast to existing work, that directly concatenate view point parameters with latent codes, we first encode the image into a latent representation which is explicitly rotated and translated in Euclidean space. 
We then decode the transformed latent code, which is assumed to implicitly represent the  3D geometry, into a depth map in target view.
From the depth map we compute dense correspondences between pixels in the source and target view via perspective projection and subsequently the final output image via pixel warping. 
All operations involved are differentiable, allowing for end-to-end training.

Detailed experiments are performed on synthetic objects \cite{chang2015shapenet} and natural images \cite{Geiger2012CVPR}. 
We assess the image quality, granularity, precision of continuous viewpoint control and implicit recovery of scene geometry qualitatively and quantitatively. 
Our experiments demonstrate that both components, the TAE and depth-guided warping, drastically improve the robustness and accuracy for continuous view synthesis.

In conclusion, our main contributions are:
\begin{compactitem}
    \item We propose the task of continuous view synthesis from monocular inputs under fine-grained view control.
    \item This goal is achieved via a proposed novel architecture that integrates a transforming encoder-decoder network and depth-guided image mapping. 
    \item Thorough experiments are conducted, demonstrating the efficacy of our method compared to prior art. 
\end{compactitem}

\section{Related Work}
\vspace{-0.2cm}
\myparagraph{View synthesis with multi-view images.}
The task of synthesizing new views given a sequence of images as input has been studied intensely in both the vision and graphics community.
Strategies can be classified into those
that explicitly compute a 3D representation of the
scene~\cite{rematas2017novel, kopf2013image,penner2017soft,debevec1996modeling,he1998layered,seitz2006comparison,zitnick2004high,chaurasia2013depth,ladicky2014pulling}, and those in which the 3D
geometry is handled implicitly~\cite{fitzgibbon2005image,matusik2002image,mcmillan1995plenoptic}. Others have deployed full 4D light fields \cite{gortler1996lumigraph, levoy1996light}, albeit at the cost of complex hardware setups and increased computational cost.
Recently, deep learning techniques have been applied in similar settings to fill holes and eliminate artifacts caused by the sampling gap, dis-occlusions, and inaccurate 3D reconstructions \cite{flynn2016deepstereo,HPPFDB18,zhou2018stereo,Wang2018-ox,Srinivasan2019-hx,Flynn2019-lk,Mildenhall2019-xk}.
While improving results over traditional methods, such approaches rely on multi-view input and are hence limited to the same setting.

\myparagraph{View synthesis with monocular input.}
Recent work leverages deep neural networks to learn a monocular image-to-image mapping between source and target view from data \cite{kulkarni2015deep,TDB16a,dosovitskiy2017learning,zhou2016view,tvsn_cvpr2017,sun2018multiview,rnn_view_synthesis}.
One line of work \cite{kulkarni2015deep,TDB16a,dosovitskiy2017learning,olszewski2019tbn} directly generates image pixels. Given the difficulty of the task, direct image-to-image translation approaches struggle with preservation of local details and often produce blurry images.
Zhou et.al. \cite{zhou2016view} estimate flow maps in order to warp source view pixels to their location in the output.
Others further refine the results by image completion \cite{tvsn_cvpr2017} or by fusing multiple views \cite{sun2018multiview}.

Typically, the desired view is controlled by concatenating latent codes with a flattened viewpoint transform. 
However, the exact mapping between viewpoint parameters to images is difficult to learn due to sparse training pairs from the continuous viewpoint space. We show experimentally that this leads to a snapping to training views, with image quality quickly degrading under continuous view control.
Recent works demonstrate the potential for fine-grained view synthesis, but either are limited to single instances of objects \cite{sitzmann2019deepvoxels} or require additional supervision in the form of depth maps \cite{zhu,liu2018geometry},
 surface normals \cite{liu2018geometry} and even light field images \cite{srinivasan2017learning}, which are cumbersome to acquire in real settings. In contrast, our method consists of a fully differentiable network, which is trained with image pairs and associated transformations as sole supervision.

\myparagraph{3D from single image.}
Reasoning about the 3D shape can serve as an implicit step of free-from view synthesis.
Given the severely under-constrained case of recovering 3D shapes from a single image, recent works have deployed neural networks for this task.
They can be categorized by their output representation into 
mesh \cite{kato2018renderer,DBLP:journals/corr/abs-1901-05567}, 
point cloud \cite{fan2017point,lin2018learning, insafutdinov18pointclouds}, 
voxel~\cite{wu20153d, choy20163d,girdhar2016learning,tulsiani2017multi,riegler2017octnet}, 
or depth map based~\cite{eigen2014depth, zhou2017unsupervised, Tulsiani_2018_ECCV}. 
Mesh-based approaches are still not accurate enough due to the indirect learning process. Point clouds are often sparse and cannot be directly leveraged to project dense color information in the output image and voxel-based methods are limited in resolution and number and type of objects due to memory constraints. Depth maps become sparse and incomplete when projected into other views due to the sampling gap and occlusions. Layered depth map representations \cite{Tulsiani_2018_ECCV} have been used to alleviate this problem. However, a large number of layers would be necessary which poses significant hurdles in terms of scalability and runtime efficiency.
In contrast to explicit models, our latent 3D geometry representation is compact and memory efficient, is meaningful under explicit 3D transformation and can be used to render dense images.

\myparagraph{Deep generative models.}
View synthesis can also be seen as an image generation process, which is related to the field of deep generative modelling of images \cite{kingma2013auto, goodfellow2014generative}.
Recent models \cite{brock2018large,karras2018style} are able to generate high-fidelity images with diversity in many aspects including viewpoint, shape and appearance, but offer little to no exact control over the underlying parameters.
Disentangling latent factors has been studied in \cite{chen2016infogan, higgins2017beta} to provide control over image attributes. In particular, recent work \cite{zhu2018visual,Nguyen-Phuoc2019-fv} demonstrates inspiring results of viewpoint disentanglement by reasoning about the geometry. Although such methods can be used for view synthesis, the generated views lack consistency and moreover one cannot control which object to synthesize.

\section{Method}

\figureNetwork

Our main contribution is a novel geometry aware network design, shown in  \figref{fig:pipeline}, that consists of four components: 3D transforming auto-encoder (TAE), self-supervised depth map prediction, depth map projection and appearance warping.

The source view is first encoded into a latent code ($ z = E_{\theta_e}(I_s)$). 
This latent code $z$ is encouraged by our learning scheme to be meaningful in 3D metric space. 
After encoding we apply the desired transformation between the source and target to the latent code. 
The transformed code ($z_T = T_{s \rightarrow t}(z)$) is decoded by a neural network to predict a depth map $D_t$ as observed from the target viewpoint. 
$D_t$ is projected back into the source view based on the known camera intrinsics $K$ and extrinsics $T_{s \rightarrow t}$, yielding dense correspondences between the target and source views, encoded as dense backward flow map $C_{t \rightarrow s}$. 
This flow map is used to warp the source view pixel-by-pixel into the target view. 

Note that attaining backward flow and hence predicting depth maps in the \emph{target} view is a crucial difference to prior work. 
Forward mapping of pixel values into the target view $I_t$ would incur discretization artifacts when moving between ray and pixel-space, visible as banding after re-projection of the (source view) depth map. 
The whole network is trained end-to-end with a simple per-pixel reconstruction loss as sole guidance. 
Overall, we want to learn a mapping $M:X\rightarrow Y$, which in our case can be decomposed as:
\begin{align}
M(I_s) = B(P_{t \rightarrow s}(D_{\theta_d}(T_{s \rightarrow t}(E_{\theta_e}(I_s)))), I_s) = \hat{I}_t,
\label{equ:mapping}
\end{align}
where $B$ is the bi-linear sampling function, $P_{t\rightarrow s}$ is the perspective projection, and $E_{\theta_e},D_{\theta_d}$ are the encoder and decoder networks respectively. 
This decomposition is an important contribution of our work. By asking the network to predict a depth map $D_t$ in the target view, we implicitly encourage the TAE  encoder $E_{\theta_e}$ to produce position predictions for features and the decoder $D_{\theta_d}$ learns to generate features at corresponding positions by rendering the transformed representation from the specified view-angle.  

\subsection{Transforming Auto-encoder} 
We take inspiration from recent work \cite{sabour2017dynamic,hinton2018matrix,Worrall2017InterpretableTW,Rhodin} which itself builds upon earlier work by Hinton et al.~\cite{Hinton:2011:TA:2029556.2029562}, that uses encoder-decoder architectures to learn representations that are transformation equivariant, establishing a direct correspondence between image and feature spaces. We leverage such a latent space to model the relationship between viewpoint and implicit 3D shape.

To this end, we represent the latent code $z_s$ as vectorized set of points $z_s\in \mathbb{R}^{n\times 3}$, where $n$ is a hyper-parameter. 
This representation is then multiplied with the ground-truth transformation $T_{s\rightarrow t} = [R|t]_{s\rightarrow t}$ describing the viewpoint change between source view $I_s$ and target view $I_t$ to attain the rotated code $z_t$:
{
\begin{align}
z_t = [R|t]_{s\rightarrow t} \cdot \Tilde{z_s},
\label{equ:transformation}
\end{align}
}
where $\Tilde z_s$ is the homogeneous representation of $z_s$. 
In this way the network is trained to encode position predictions for features which can then be decoded into images.  All functions in the TAE module including encoding, vector reshaping, matrix multiplication and decoding are differentiable and hence amenable to training via backpropagation. \\

\vspace{-0.2cm}
\subsection{Depth Guided Appearance Mapping}

We decode $z_t$ into 3D shape in the target view, represented as a depth image $D_t$. 
From $D_t$ we compute the dense correspondence field $C_{t\rightarrow s}$ deterministically via perspective projection $P_{t \rightarrow s}$. 
The dense correspondences are then used to warp the pixels of the texture (source view) $I_s$ into the target view $\hat{I}_t$. 
This allows the network to warp the source view into the target view and makes the prediction of target view invariant to the texture of the input, resulting in sharp and detail-preserving outputs. 

\myparagraph{Establishing correspondences.}
The per-pixel correspondences $C_{t\rightarrow s}$ are attained from the depth image $D_t$ in the target view by conversion from the depth map to 3D coordinates $[X,Y,Z]$ and perspective projection:

{\footnotesize
\begin{align}
[X,Y,Z]^T = D_t(x_t,y_t)K^{-1}[x_t, y_t,1]^T \quad \\
\text{and} \quad
[x_s,y_s,1]^T \sim KT_{t\rightarrow s}[X,Y,Z,1]^T \quad .
\label{equ:projection}
\end{align}
}
where each pixel $(x_t,y_t)$ encodes the corresponding pixel position in the source view $(x_s,y_s)$. Furthermore, $K$ is the camera intrinsic matrix describing normalized focal length along both axes $f_x, f_y$ and image center $c_x, c_y$. 
Note that only the focal length ratio $f_x/f_y$ as well as image center affect view synthesis, while the absolute scale of the focal length is only important to predict geometry at correct scale.
\\

\myparagraph{Warping with correspondences.}
With the dense correspondences obtained, we are now able to warp the source view to the target view. 
This operation propagates texture and local details. 
Since the corresponding pixel positions that are derived from Eq.~\ref{equ:projection} are non-integer, this is done via differentiable bilinear sampling as proposed in \cite{jaderberg2015spatial}:

{\footnotesize
\begin{align}
\begin{split}
I_{t}(x_t,y_t) = 
\sum_{x_s} \sum_{y_s} I_s(x_s,y_s)
\text{max}(0, 1-| x_s - C_{x}(x_t,y_t)|)\\ \text{max}(0, 1-|y_s - C_{y}(x_t,y_t)|) .
\end{split}
\end{align}
}%

The use of backward flow $C_{t \rightarrow s}$, computed from the predicted depth map $D_t$, makes the approach amenable to gradient based optimization since the gradient of the per-pixel reconstruction loss provides meaningful information to correct erroneous correspondences. 
The gradients also flow back to provide useful information to the TAE network owing to the fact that the correspondences are computed deterministically from the predicted depth maps. 
While bearing similarity to \cite{zhou2016view}, we introduce the intermediate step of predicting depth, instead of predicting the correspondences directly. 
This enforces the network to obey geometric constraints, resolving ambiguous correspondences.

\subsection{Training}\label{sec:training}
All steps in our network, namely 3D transforming auto-encoder (TAE), self-supervised depth map prediction, depth map projection and appearance warping, are differentiable which enables end-to-end training. 
Among all modules, only the TAE module contains trainable parameters ($\theta_e,\theta_d$). 
To train the network only pairs of source and target views and their transformation are required.  
The network weights are optimized via minimization of the $L_1$ loss between the predicted target view $\hat{I_t}$ and the ground truth $I_t$. 

{\footnotesize
\begin{align}
\mathcal{L}_{recon} = \norm{I_t- \hat{I}_{t}}_1
\end{align}
}

Minimizing this reconstruction loss, the network learns to produce realistic novel views, to predict the necessary flow and depth maps and learn to form a geometrical latent space. 
\section{Experiments}

\figureShapeNet

We now evaluate our method quantitatively and qualitatively. We are especially interested in assessing image quality, granularity and precision of fine-grained viewpoint control. First, we conduct detailed experiments on synthetic objects, where ground-truth of continuous viewpoint is easy to obtain, to numerically assess the reconstruction quality. Notably, we vary the viewpoints in much smaller step-sizes than what is observed in the training data. 
Second, to evaluate generalizability, we test our system on natural city scenes. In this setting, given an image input, we specify the desired ground-truth camera trajectories along which the system generates novel views. Then we run an existing visual odometry system on these synthesized continuous views to recover the camera trajectory. By comparing the recovered trajectory with the ground-truth, we can evaluate the geometrical property of the synthesized images under the consideration of granularity and continuous view control.
Finally, to better understand the mechanism of our proposed network, we further conduct studies on its two key components, namely depth-guided texture mapping and transforming auto-encoder. 
We evaluate the intermediate depth and flow, and qualitatively verify the meaningfulness of the latent space of the TAE.

\subsection{Datasets}
\vspace{-0.1cm}
\noindent We conduct our experiments on two challenging datasets: synthetic objects~\cite{chang2015shapenet} and real natural scenes~\cite{Geiger2012CVPR}. 

\myparagraph{ShapeNet \cite{chang2015shapenet}}
is a large collection of 3D synthetic objects from various categories. 
Similar to \cite{zhou2016view,tvsn_cvpr2017,sun2018multiview} we choose \textbf{car} and \textbf{chair} to evaluate our method. 
We use the same train test split as proposed in \cite{zhou2016view}. 
For training we render each models from 54 viewpoints with different azimuth and elevation. The azimuth goes from $0^{\circ}$ to $360^{\circ}$ with a step size of $20^{\circ}$ and the elevation from $0^{\circ}$ to $30^{\circ}$ with a step size of $10^{\circ}$. 
Each training pair consists of two views of the same instance, with a difference in azimuth within $\pm 40^{\circ}$.

\myparagraph{KITTI \cite{Geiger2012CVPR}}
is a standard dataset for autonomous driving, containing complex city scenes in uncontrolled environments. 
We conduct experiments on the KITTI odometry subset which contains image sequences as well as the global camera poses of each frame. 
In total there are 18560 images for training and 4641 images for testing. 
We construct training pairs by randomly selecting target view among 10 nearest frames of source view. 
The relative transformation is obtained from the global camera poses.

\subsection{Metrics}
\vspace{-0.1cm}
\noindent In our evaluations we report the following metrics:

\myparagraph{Mean Absolute Error $L_1$} is used to measure per-pixel value differences between ground-truth and the predictions. 

\myparagraph{Structural SIMilarity (SSIM) Index}\cite{wang2004image} has values in [-1, 1] and measures the structural similarity between synthesized image and ground truth. We report SSIM in addition to the $L_1$ loss since it i) gives an indication of perceptual image quality and ii) serves as further metric that is not directly optimized during training.

\myparagraph{Percentage of correctness under threshold $\delta$ (Acc)}. The predicted flow/depth $\hat{y_i}$ at pixel $i$, given ground truth $y_i$, is regarded as correct if $max(\frac{y_i}{\hat{y_i}},\frac{\hat{y_i}}{y_i}) < \delta$ is satisfied. We count the portion of correctly predicted pixels. Here $\delta=1.05$.

\myparagraph{Rotation error and translation error}
are defined as:
{\footnotesize
\begin{align}
&RE = arccos(\frac{\mathbf{Tr}(\Tilde{R} \cdot R^T )-1}{2}), TE = arccos( \frac{\Tilde{t} \cdot t^T}{ \Tilde{\norm{t}}_2 \cdot \norm{t}_2 })
\end{align}
}%
where $\mathbf{Tr}$ represents the trace of the matrix.

\subsection{Comparison with other methods}
\vspace{-0.1cm}
We compare with several representative state-of-the-art learning-based view synthesis methods. 
\myparagraph{Tatarchenko et al.}~\cite{TDB16a}
treat the view synthesis as an image-to-image translation task and generate pixels directly. In their framework the viewpoint is directly concatenated with the latent code. 
\myparagraph{Zhou et al.}~\cite{zhou2016view}
generates flow instead of pixels. The view information is also directly concatenated. 
\myparagraph{Sun et al.}~\cite{sun2018multiview}
combines both pixel generation \cite{TDB16a} and image warping \cite{zhou2016view}. 
The original implementation in Zhou et al.~\cite{zhou2016view} and Sun et al.~\cite{sun2018multiview} does not support continuous viewpoint input for objects. To allow for continuous input for comparison, we replace their encoded discrete one hot viewpoint representation with cosine and sine values of the view angles. The same encoder and decoder are used for all comparisons.

\figureControl

\subsection{ShapeNet Evaluation}
\vspace{-0.1cm}
\figureResultKITTIImage

To test the granularity and precision of viewpoint control, for each test object, given a source view $I_s$, the network synthesizes 80 views around the source view with a step size of $1^\circ$ which is much denser than the step size of $20^\circ$ for training (and much denser than previously reported experiments). In total the test set contains 100,000 view pairs of objects.

To study the effectiveness of the transformation-aware latent space, we introduce \textbf{Ours (w/o TAE)} concatenating the viewpoint analogously to \cite{TDB16a,dosovitskiy2017learning,zhou2016view,tvsn_cvpr2017,zhou2016view} while still keeping the depth-guided texture mapping process. To evaluate the depth-guided texture mapping process, we introduce \textbf{Ours (w/o depth)} which directly predicts flow without the depth guidance but does deploy the TAE.

\myparagraph{Viewpoint dependent error.}
Fig.~\ref{fig:control} plots the L1 reconstruction error between $[-40^\circ,40^\circ]$ of all methods. Note that $0^\circ$ here means no transformation applied to the source view. 
\emph{Ours} consistently produces lower errors. More importantly it yields much lower variance between non-training and training views ($\pm40^\circ,\pm20^\circ$ are training views). 
While previous methods can achieve similar performance to ours at training views, their performance significantly decreases for non-training views. 
Notably, both of our designs (TAE and depth-based appearance) contribute to the final performance and the problem of snapping to training views persists with either of the two components discarded (\textbf{Ours (w/o TAE)} and \textbf{Ours (w/o depth)}). 
Tab.~\ref{tab:shapenet_view} summarizes the average $L_1$ error and SSIM for all generated views between $[-40^\circ,40^\circ]$. 
Inline with Fig.~\ref{fig:control}, our method significantly outperforms previous methods on both car and chair. 
In addition, both of our ablative methods also perform better than previous methods, demonstrating the effectiveness of both modules. 
\tableShapeNetView

\myparagraph{Qualitative results.}
The qualitative results in Fig.~\ref{fig:large_shapenet} confirm the quantitative findings. 
To demonstrate the capability of continuous viewpoint control, we generate and overlay 80 views with step size of $1^\circ$ from a single input.
Compared to previous approaches, our method exhibits similar spin pattern as the ground truth, whereas other methods mostly snap to the fixed training views (Zhou et al.~\cite{zhou2016view}, Sun et al.~\cite{sun2018multiview}). 
This suggests that overfitting occurs, limiting the granularity and precision of view control. A close look at specific views reveals that previous methods display distortions at non-training views, highlighted in red. 
The image generated by Tatarchenko et al.~\cite{TDB16a} is blurry.

\figureResultKITTISLAM
\subsection{KITTI Evaluation}%
\vspace{-0.1cm}
We now evaluate our method in the more realistic setting of the KITTI dataset. Note that the dataset only contains fairly linear forward motion recorded from a car's dash. This setting is a good testbed for the envisioned application scenarios where one desires to extract 3D information retroactively.

\myparagraph{Qualitative results}
In \figref{fig:large_kitti2} we show qualitative results from novel views synthesized along a straight camera trajectory:
Zhou et al.~\cite{zhou2016view} and Sun et al.~\cite{sun2018multiview} both have difficulties to deal with viewpoints outside of the training setting and produce distorted images while ours are sharp and geometrically correct. Ours more faithfully reproduces the desired motion than \cite{zhou2016view} and \cite{sun2018multiview} which remains stationary. 

\myparagraph{Complex trajectory recovery.}
To simulate real use cases, we introduce a new experimental setting. We specify arbitrary \textit{desired} trajectories, specifically so that the camera moves away from the car's original motion. From this specification we generate a sequences of 100 images along the trajectories. 
Subsequently we run a state-of-the-art visual odometry \cite{engel2017direct} system to estimate the camera pose based on the \emph{synthesized} views. If the view synthesis approach is geometrically accurate, the visual odometry system should recover the \textit{desired} trajectory. 
\figref{fig:large_kitti} illustrates one such experiment. 
The estimated trajectory from ours aligns well with the ground-truth. 
In contrast, views from \cite{zhou2016view} result in a wrong trajectory and \cite{sun2018multiview} mostly produce straight forward motion, possibly due to overfitting to training trajectories. 
\myparagraph{Quantitative results.}
To evaluate the geometrical properties quantitatively, we generate new views with randomly sampled transformation $T=[R|t]$. 
We then estimate the relative transformation between the input and the synthesized view $\Tilde{T}=[\Tilde{R}|\Tilde{t}]$ and compare to the ground-truth $T$. 
This is done by first detecting and matching SURF features \cite{bay2006surf} in both views, and then computing and decomposing the essential matrix. 
We report the numerical error in Tab.~\ref{tab:kitti_view}. 
Our method produces drastically lower error in rotation and the translation, indicating accurate viewpoint control. 
Note that we had to remove \cite{TDB16a} from this comparison since SURF feature detection fails due to the very blurry images. 
\tableKITTIView

\vspace{-0.3cm}
\subsection{Depth and Flow Evaluation}
\vspace{-0.1cm}
The quality of predicted depth map and warping flow is essential to produce geometrically correct views. 
We evaluate the accuracy of depth and flow prediction with two metrics ($L_1$ and Acc).
Tab.~\ref{tab:depth_flow} summarizes results for ShapeNet. 
Ours achieves the best accuracy in both flow and depth prediction, which directly benefits view synthesis (cf. Tab.~\ref{tab:shapenet_view}). 
The relative ranking of the ablative baselines furthermore indicates that both the TAE and the depth-guided texture mapping help to improve the flow accuracy. The TAE furthermore guides the depth prediction. 
To illustrate that the reconstructed depth maps are indeed meaningful, we predict depth in different target views and visualize the extracted normal maps, as shown in Fig.~\ref{fig:pcl}. 

\myparagraph{Discussion}
Together these experiments indicate that the proposed self-supervision indeed forces the network to infer underlying 3D structure (yielding good depth which is necessary for accurate flow maps) and that it helps the final task without requiring additional labels. 

\tableDepthFlow
\vspace{-0.3cm}
\figurePCL

\vspace{-0.3cm}
\subsection{Latent Space Analysis}
\vspace{-0.1cm}
To verify that the learned latent space is indeed interpretable and meaningful under geometrical transformation, we i) linearly interpolate between latent points of two objects and ii) rotate each interpolated latent point set. These point sets are then decoded into depth maps, visualized as normal maps in the global frame. Fig.~\ref{fig:inter} shows that interpolated samples exhibit a smooth shape transition while the viewpoint remains constant (i). Moreover, rotating the latent points only changes the viewpoint without affecting the shape (ii). 
\vspace{-0.2cm}
\figureInterpolation
\vspace{-0.5cm}
\subsection{Generalization to unseen data}
\vspace{-0.1cm}
We find that our model generalizes well to unseen data thanks to the usage of depth-based warping. 
Interestingly, our model trained on $256^2$ images can be directly applied to high resolution ($1024^2$) images without additional training. 
The inference process takes 50ms per frame on a Titan X GPU, allowing for real time rendering of synthetized views. 
This enables many appealing application scenarios. 
For example, our model, trained on ShapeNet only, can be used in an app where downloaded 2D images are brought to life and a user may browse the depicted object in 3D. 
With a model trained on KITTI, a user may explore a 3D scene from a single image, via generation of free-viewpoint videos or AR/VR content (see \figref{fig:teaser}). 
\vspace{-0.2cm}
\section{Conclusion}
\vspace{-0.1cm}
We have presented a novel learning pipeline for continuous view synthesis. 
At its core lies a depth-based image prediction network that is forced to satisfy explicitly formulated geometric constraints. 
The latent representation is meaningful under explicit 3D transformation and can be used to produce geometrically accurate views for both single objects and natural scenes. 
We have conducted thorough experiments on synthetic and natural images and have demonstrated the efficacy of our approach.

{
\noindent{\bf Acknowledgement.}
We thank Nvidia for the donation of GPUs used in this work. We would like to express our gratitude to Olivier Saurer, Velko Vechev, Manuel Kaufmann, Adrian Spurr, Yinhao Huang, Xucong Zhang and David Lindlbauer for the insightful discussions, James Bern and Seonwook Park for the video voice-over.
}

{\small
\bibliographystyle{misc/ieee_fullname}
\bibliography{ref}

\begin{thebibliography}{10}\itemsep=-1pt

\bibitem{bay2006surf}
Herbert Bay, Tinne Tuytelaars, and Luc Van~Gool.
\newblock Surf: Speeded up robust features.
\newblock In {\em Proc. of the European Conf. on Computer Vision (ECCV)}, 2006.

\bibitem{brock2018large}
Andrew Brock, Jeff Donahue, and Karen Simonyan.
\newblock Large scale gan training for high fidelity natural image synthesis.
\newblock In {\em Proc. of the International Conf. on Learning Representations
  (ICLR)}, 2018.

\bibitem{chang2015shapenet}
Angel~X Chang, Thomas Funkhouser, Leonidas Guibas, Pat Hanrahan, Qixing Huang,
  Zimo Li, Silvio Savarese, Manolis Savva, Shuran Song, Hao Su, et~al.
\newblock Shapenet: An information-rich 3d model repository.
\newblock {\em arXiv preprint arXiv:1512.03012}, 2015.

\bibitem{chaurasia2013depth}
Gaurav Chaurasia, Sylvain Duchene, Olga Sorkine-Hornung, and George Drettakis.
\newblock Depth synthesis and local warps for plausible image-based navigation.
\newblock {\em ACM Transactions on Graphics (TOG)}, 32(3):30, 2013.

\bibitem{chen2016infogan}
Xi Chen, Yan Duan, Rein Houthooft, John Schulman, Ilya Sutskever, and Pieter
  Abbeel.
\newblock Infogan: Interpretable representation learning by information
  maximizing generative adversarial nets.
\newblock In {\em Advances in Neural Information Processing Systems (NIPS)},
  2016.

\bibitem{choy20163d}
Christopher~B Choy, Danfei Xu, JunYoung Gwak, Kevin Chen, and Silvio Savarese.
\newblock 3d-r2n2: A unified approach for single and multi-view 3d object
  reconstruction.
\newblock In {\em Proc. of the European Conf. on Computer Vision (ECCV)}, 2016.

\bibitem{debevec1996modeling}
Paul~E Debevec, Camillo~J Taylor, and Jitendra Malik.
\newblock Modeling and rendering architecture from photographs: A hybrid
  geometry-and image-based approach.
\newblock In {\em Conference on Computer Graphics and Interactive Techniques},
  pages 11--20. ACM, 1996.

\bibitem{dosovitskiy2017learning}
Alexey Dosovitskiy, Jost~Tobias Springenberg, Maxim Tatarchenko, and Thomas
  Brox.
\newblock Learning to generate chairs, tables and cars with convolutional
  networks.
\newblock {\em IEEE transactions on pattern analysis and machine intelligence},
  39(4):692--705, 2016.

\bibitem{eigen2014depth}
David Eigen, Christian Puhrsch, and Rob Fergus.
\newblock Depth map prediction from a single image using a multi-scale deep
  network.
\newblock In {\em Advances in Neural Information Processing Systems (NIPS)},
  2014.

\bibitem{engel2017direct}
Jakob Engel, Vladlen Koltun, and Daniel Cremers.
\newblock Direct sparse odometry.
\newblock {\em IEEE transactions on pattern analysis and machine intelligence},
  40(3):611--625, 2017.

\bibitem{fan2017point}
Haoqiang Fan, Hao Su, and Leonidas~J Guibas.
\newblock A point set generation network for 3d object reconstruction from a
  single image.
\newblock In {\em Proc. IEEE Conf. on Computer Vision and Pattern Recognition
  (CVPR)}, 2017.

\bibitem{fitzgibbon2005image}
Andrew Fitzgibbon, Yonatan Wexler, and Andrew Zisserman.
\newblock Image-based rendering using image-based priors.
\newblock {\em International Journal of Computer Vision}, 63(2):141--151, 2005.

\bibitem{Flynn2019-lk}
John Flynn, Michael Broxton, Paul Debevec, Matthew DuVall, Graham Fyffe, Ryan
  Overbeck, Noah Snavely, and Richard Tucker.
\newblock Deepview: View synthesis with learned gradient descent.
\newblock In {\em Proc. IEEE Conf. on Computer Vision and Pattern Recognition
  (CVPR)}, 2019.

\bibitem{flynn2016deepstereo}
John Flynn, Ivan Neulander, James Philbin, and Noah Snavely.
\newblock Deepstereo: Learning to predict new views from the world's imagery.
\newblock In {\em Proc. IEEE Conf. on Computer Vision and Pattern Recognition
  (CVPR)}, 2016.

\bibitem{Geiger2012CVPR}
Andreas Geiger, Philip Lenz, and Raquel Urtasun.
\newblock Are we ready autonomous driving? the kitti vision benchmark suite.
\newblock In {\em Proc. IEEE Conf. on Computer Vision and Pattern Recognition
  (CVPR)}, 2012.

\bibitem{girdhar2016learning}
Rohit Girdhar, David~F Fouhey, Mikel Rodriguez, and Abhinav Gupta.
\newblock Learning a predictable and generative vector representation for
  objects.
\newblock In {\em Proc. of the European Conf. on Computer Vision (ECCV)}, 2016.

\bibitem{goodfellow2014generative}
Ian Goodfellow, Jean Pouget-Abadie, Mehdi Mirza, Bing Xu, David Warde-Farley,
  Sherjil Ozair, Aaron Courville, and Yoshua Bengio.
\newblock Generative adversarial nets.
\newblock In {\em Advances in Neural Information Processing Systems (NIPS)},
  2014.

\bibitem{gortler1996lumigraph}
Steven~J Gortler, Radek Grzeszczuk, Richard Szeliski, and Michael~F Cohen.
\newblock The lumigraph.
\newblock In {\em Conference on Computer Graphics and Interactive Techniques},
  volume~96, pages 43--54, 1996.

\bibitem{HPPFDB18}
Peter Hedman, Julien Philip, True Price, Jan-Michael Frahm, George Drettakis,
  and Gabriel Brostow.
\newblock Deep blending for free-viewpoint image-based rendering.
\newblock In {\em SIGGRAPH Asia 2018 Technical Papers}, page 257. ACM, 2018.

\bibitem{higgins2017beta}
Irina Higgins, Loic Matthey, Arka Pal, Christopher Burgess, Xavier Glorot,
  Matthew Botvinick, Shakir Mohamed, and Alexander Lerchner.
\newblock beta-vae: Learning basic visual concepts with a constrained
  variational framework.
\newblock In {\em Proc. of the International Conf. on Learning Representations
  (ICLR)}, 2017.

\bibitem{Hinton:2011:TA:2029556.2029562}
Geoffrey~E. Hinton, Alex Krizhevsky, and Sida~D. Wang.
\newblock Transforming auto-encoders.
\newblock In {\em International Conference on Artificial Neural Networks},
  2011.

\bibitem{hinton2018matrix}
Geoffrey~E Hinton, Sara Sabour, and Nicholas Frosst.
\newblock Matrix capsules with em routing.
\newblock In {\em Proc. of the International Conf. on Learning Representations
  (ICLR)}, 2018.

\bibitem{insafutdinov18pointclouds}
Eldar Insafutdinov and Alexey Dosovitskiy.
\newblock Unsupervised learning of shape and pose with differentiable point
  clouds.
\newblock In {\em Advances in Neural Information Processing Systems (NIPS)},
  2018.

\bibitem{jaderberg2015spatial}
Max Jaderberg, Karen Simonyan, Andrew Zisserman, et~al.
\newblock Spatial transformer networks.
\newblock In {\em Advances in Neural Information Processing Systems (NIPS)},
  2015.

\bibitem{karras2018style}
Tero Karras, Samuli Laine, and Timo Aila.
\newblock A style-based generator architecture for generative adversarial
  networks.
\newblock In {\em Proc. IEEE Conf. on Computer Vision and Pattern Recognition
  (CVPR)}, 2019.

\bibitem{kato2018renderer}
Hiroharu Kato, Yoshitaka Ushiku, and Tatsuya Harada.
\newblock Neural 3d mesh renderer.
\newblock In {\em Proc. IEEE Conf. on Computer Vision and Pattern Recognition
  (CVPR)}, 2018.

\bibitem{kingma2013auto}
Diederik~P Kingma and Max Welling.
\newblock Auto-encoding variational bayes.
\newblock In {\em Proc. of the International Conf. on Learning Representations
  (ICLR)}, 2013.

\bibitem{kopf2013image}
Johannes Kopf, Fabian Langguth, Daniel Scharstein, Richard Szeliski, and
  Michael Goesele.
\newblock Image-based rendering in the gradient domain.
\newblock {\em ACM Transactions on Graphics (TOG)}, 32(6):199, 2013.

\bibitem{kulkarni2015deep}
Tejas~D Kulkarni, William~F Whitney, Pushmeet Kohli, and Josh Tenenbaum.
\newblock Deep convolutional inverse graphics network.
\newblock In {\em Advances in Neural Information Processing Systems (NIPS)},
  2015.

\bibitem{ladicky2014pulling}
Lubor Ladicky, Jianbo Shi, and Marc Pollefeys.
\newblock Pulling things out of perspective.
\newblock In {\em Proc. IEEE Conf. on Computer Vision and Pattern Recognition
  (CVPR)}, 2014.

\bibitem{levoy1996light}
Marc Levoy and Pat Hanrahan.
\newblock Light field rendering.
\newblock In {\em Conference on Computer Graphics and Interactive Techniques},
  pages 31--42. ACM, 1996.

\bibitem{lin2018learning}
Chen-Hsuan Lin, Chen Kong, and Simon Lucey.
\newblock Learning efficient point cloud generation for dense 3d object
  reconstruction.
\newblock In {\em Thirty-Second AAAI Conference on Artificial Intelligence},
  2018.

\bibitem{liu2018geometry}
Miaomiao Liu, Xuming He, and Mathieu Salzmann.
\newblock Geometry-aware deep network for single-image novel view synthesis.
\newblock In {\em Proc. IEEE Conf. on Computer Vision and Pattern Recognition
  (CVPR)}, 2018.

\bibitem{DBLP:journals/corr/abs-1901-05567}
Shichen Liu, Weikai Chen, Tianye Li, and Hao Li.
\newblock Soft rasterizer: Differentiable rendering for unsupervised
  single-view mesh reconstruction.
\newblock {\em arXiv preprint arXiv:1901.05567}, 2019.

\bibitem{matusik2002image}
Wojciech Matusik, Hanspeter Pfister, Addy Ngan, Paul Beardsley, Remo Ziegler,
  and Leonard McMillan.
\newblock Image-based 3d photography using opacity hulls.
\newblock In {\em ACM Transactions on Graphics (TOG)}, volume~21, pages
  427--437. ACM, 2002.

\bibitem{mcmillan1995plenoptic}
Leonard McMillan and Gary Bishop.
\newblock Plenoptic modeling: An image-based rendering system.
\newblock In {\em Conference on Computer Graphics and Interactive Techniques},
  pages 39--46. Citeseer, 1995.

\bibitem{Mildenhall2019-xk}
Ben Mildenhall, Pratul~P. Srinivasan, Rodrigo Ortiz-Cayon, Nima~Khademi
  Kalantari, Ravi Ramamoorthi, Ren Ng, and Abhishek Kar.
\newblock Local light field fusion: Practical view synthesis with prescriptive
  sampling guidelines.
\newblock {\em ACM Transactions on Graphics (TOG)}, 2019.

\bibitem{Nguyen-Phuoc2019-fv}
Thu Nguyen-Phuoc, Chuan Li, Lucas Theis, Christian Richardt, and Yong-Liang
  Yang.
\newblock Hologan: Unsupervised learning of 3d representations from natural
  images.
\newblock {\em arXiv preprint arXiv:1904.01326}, 2019.

\bibitem{olszewski2019tbn}
Kyle Olszewski, Sergey Tulyakov, Oliver Woodford, Hao Li, and Linjie Luo.
\newblock Transformable bottleneck networks.
\newblock {\em arXiv:1904.06458}, 2019.

\bibitem{tvsn_cvpr2017}
Eunbyung Park, Jimei Yang, Ersin Yumer, Duygu Ceylan, and Alexander~C Berg.
\newblock Transformation-grounded image generation network for novel 3d view
  synthesis.
\newblock In {\em Proc. IEEE Conf. on Computer Vision and Pattern Recognition
  (CVPR)}, 2017.

\bibitem{penner2017soft}
Eric Penner and Li Zhang.
\newblock Soft 3d reconstruction for view synthesis.
\newblock {\em ACM Transactions on Graphics (TOG)}, 36(6):235, 2017.

\bibitem{rematas2017novel}
Konstantinos Rematas, Chuong~H Nguyen, Tobias Ritschel, Mario Fritz, and Tinne
  Tuytelaars.
\newblock Novel views of objects from a single image.
\newblock {\em IEEE transactions on pattern analysis and machine intelligence},
  39(8):1576--1590, 2016.

\bibitem{Rhodin}
Helge Rhodin, Mathieu Salzmann, and Pascal Fua.
\newblock Unsupervised geometry-aware representation for 3d human pose
  estimation.
\newblock In {\em Proc. of the European Conf. on Computer Vision (ECCV)}, 2018.

\bibitem{riegler2017octnet}
Gernot Riegler, Ali Osman~Ulusoy, and Andreas Geiger.
\newblock Octnet: Learning deep 3d representations at high resolutions.
\newblock In {\em Proc. IEEE Conf. on Computer Vision and Pattern Recognition
  (CVPR)}, 2017.

\bibitem{sabour2017dynamic}
Sara Sabour, Nicholas Frosst, and Geoffrey~E Hinton.
\newblock Dynamic routing between capsules.
\newblock In {\em Advances in Neural Information Processing Systems (NIPS)},
  2017.

\bibitem{seitz2006comparison}
Steven~M Seitz, Brian Curless, James Diebel, Daniel Scharstein, and Richard
  Szeliski.
\newblock A comparison and evaluation of multi-view stereo reconstruction
  algorithms.
\newblock In {\em Computer vision and pattern recognition, 2006 IEEE Computer
  Society Conference on}, volume~1, pages 519--528. IEEE, 2006.

\bibitem{he1998layered}
Jonathan Shade, Steven Gortler, Li-wei He, and Richard Szeliski.
\newblock Layered depth images.
\newblock In {\em Conference on Computer Graphics and Interactive Techniques},
  SIGGRAPH '98, pages 231--242, New York, NY, USA, 1998. ACM.

\bibitem{sitzmann2019deepvoxels}
Vincent Sitzmann, Justus Thies, Felix Heide, Matthias Nie{\ss}ner, Gordon
  Wetzstein, and Michael Zollhofer.
\newblock Deepvoxels: Learning persistent 3d feature embeddings.
\newblock In {\em Proc. IEEE Conf. on Computer Vision and Pattern Recognition
  (CVPR)}, 2019.

\bibitem{Srinivasan2019-hx}
Pratul~P Srinivasan, Richard Tucker, Jonathan~T Barron, Ravi Ramamoorthi, Ren
  Ng, and Noah Snavely.
\newblock Pushing the boundaries of view extrapolation with multiplane images.
\newblock In {\em Proc. IEEE Conf. on Computer Vision and Pattern Recognition
  (CVPR)}, 2019.

\bibitem{srinivasan2017learning}
Pratul~P Srinivasan, Tongzhou Wang, Ashwin Sreelal, Ravi Ramamoorthi, and Ren
  Ng.
\newblock Learning to synthesize a 4d rgbd light field from a single image.
\newblock In {\em Proc. of the IEEE International Conf. on Computer Vision
  (ICCV)}, 2017.

\bibitem{sun2018multiview}
Shao-Hua Sun, Minyoung Huh, Yuan-Hong Liao, Ning Zhang, and Joseph~J Lim.
\newblock Multi-view to novel view: Synthesizing novel views with self-learned
  confidence.
\newblock In {\em Proc. of the European Conf. on Computer Vision (ECCV)}, 2018.

\bibitem{TDB16a}
Maxim Tatarchenko, Alexey Dosovitskiy, and Thomas Brox.
\newblock Multi-view 3d models from single images with a convolutional network.
\newblock In {\em Proc. of the European Conf. on Computer Vision (ECCV)}.
  Springer, 2016.

\bibitem{Tulsiani_2018_ECCV}
Shubham Tulsiani, Richard Tucker, and Noah Snavely.
\newblock Layer-structured 3d scene inference via view synthesis.
\newblock In {\em Proc. of the European Conf. on Computer Vision (ECCV)}, 2018.

\bibitem{tulsiani2017multi}
Shubham Tulsiani, Tinghui Zhou, Alexei~A Efros, and Jitendra Malik.
\newblock Multi-view supervision for single-view reconstruction via
  differentiable ray consistency.
\newblock In {\em Proc. IEEE Conf. on Computer Vision and Pattern Recognition
  (CVPR)}, 2017.

\bibitem{Wang2018-ox}
Yunlong Wang, Fei Liu, Zilei Wang, Guangqi Hou, Zhenan Sun, and Tieniu Tan.
\newblock End-to-end view synthesis for light field imaging with pseudo
  {4DCNN}.
\newblock In {\em Proceedings of the European Conference on Computer Vision
  ({ECCV})}, pages 333--348, 2018.

\bibitem{wang2004image}
Zhou Wang, Alan~C Bovik, Hamid~R Sheikh, Eero~P Simoncelli, et~al.
\newblock Image quality assessment: from error visibility to structural
  similarity.
\newblock {\em IEEE transactions on image processing}, 13(4):600--612, 2004.

\bibitem{Worrall2017InterpretableTW}
Daniel~E Worrall, Stephan~J Garbin, Daniyar Turmukhambetov, and Gabriel~J
  Brostow.
\newblock Interpretable transformations with encoder-decoder networks.
\newblock In {\em Proc. of the IEEE International Conf. on Computer Vision
  (ICCV)}, 2017.

\bibitem{wu20153d}
Zhirong Wu, Shuran Song, Aditya Khosla, Fisher Yu, Linguang Zhang, Xiaoou Tang,
  and Jianxiong Xiao.
\newblock 3d shapenets: A deep representation for volumetric shapes.
\newblock In {\em Proc. IEEE Conf. on Computer Vision and Pattern Recognition
  (CVPR)}, 2015.

\bibitem{rnn_view_synthesis}
Jimei Yang, Scott~E Reed, Ming-Hsuan Yang, and Honglak Lee.
\newblock Weakly-supervised disentangling with recurrent transformations for 3d
  view synthesis.
\newblock In {\em Advances in Neural Information Processing Systems (NIPS)},
  2015.

\bibitem{zhou2017unsupervised}
Tinghui Zhou, Matthew Brown, Noah Snavely, and David~G Lowe.
\newblock Unsupervised learning of depth and ego-motion from video.
\newblock In {\em Proc. IEEE Conf. on Computer Vision and Pattern Recognition
  (CVPR)}, 2017.

\bibitem{zhou2018stereo}
Tinghui Zhou, Richard Tucker, John Flynn, Graham Fyffe, and Noah Snavely.
\newblock Stereo magnification: Learning view synthesis using multiplane
  images.
\newblock In {\em Conference on Computer Graphics and Interactive Techniques},
  2018.

\bibitem{zhou2016view}
Tinghui Zhou, Shubham Tulsiani, Weilun Sun, Jitendra Malik, and Alexei~A Efros.
\newblock View synthesis by appearance flow.
\newblock In {\em Proc. of the European Conf. on Computer Vision (ECCV)}, 2016.

\bibitem{zhu}
Hao Zhu, Hao Su, Peng Wang, Xun Cao, and Ruigang Yang.
\newblock View extrapolation of human body from a single image.
\newblock In {\em Proc. IEEE Conf. on Computer Vision and Pattern Recognition
  (CVPR)}, 2018.

\bibitem{zhu2018visual}
Jun-Yan Zhu, Zhoutong Zhang, Chengkai Zhang, Jiajun Wu, Antonio Torralba, Josh
  Tenenbaum, and Bill Freeman.
\newblock Visual object networks: image generation with disentangled 3d
  representations.
\newblock In {\em Advances in Neural Information Processing Systems (NIPS)},
  2018.

\bibitem{zitnick2004high}
C~Lawrence Zitnick, Sing~Bing Kang, Matthew Uyttendaele, Simon Winder, and
  Richard Szeliski.
\newblock High-quality video view interpolation using a layered representation.
\newblock In {\em ACM transactions on graphics (TOG)}, volume~23, pages
  600--608. ACM, 2004.

\end{thebibliography}
}

\end{document}